\documentclass[10pt,twocolumn,letterpaper]{article}

\usepackage[pagenumbers]{wacv} % To force page numbers, e.g. for an arXiv version

\usepackage{graphicx}
\usepackage{amsmath}
\usepackage{amssymb}
\usepackage{booktabs}
\usepackage{listings}
\usepackage{xcolor}
\usepackage{float}
\usepackage{caption}
\usepackage[markup=underlined]{changes}

\usepackage[hang,flushmargin]{footmisc}

\usepackage{rotating}

\definechangesauthor[color=darkblue]{JK}
\definechangesauthor[color=darkorange]{SS}
\definecolor{darkblue}{rgb}{0.1,0.26,0.65}
\definecolor{darkorange}{rgb}{0.7,0.23,0.1}

\usepackage[pagebackref,breaklinks,colorlinks]{hyperref}
\usepackage[accsupp]{axessibility}

\usepackage[capitalize]{cleveref}
\crefname{section}{Sec.}{Secs.}
\Crefname{section}{Section}{Sections}
\Crefname{table}{Table}{Tables}
\crefname{table}{Tab.}{Tabs.}

\def\wacvPaperID{*****} % *** Enter the WACV Paper ID here
\def\confName{WACV}
\def\confYear{2024}

\begin{document}

\title{Labeling Indoor Scenes with Fusion of Out-of-the-Box Perception Models}
%\title{AutoAnnotation of Indoors Scenes for Robot Navigation}

\author{Yimeng Li$^{1}$\thanks{Equal contribution} , Navid Rajabi$^{1}$\footnotemark[1] , Sulabh Shrestha$^{1}$, Md Alimoor Reza$^{2}$,  and Jana Ko{\v{s}}eck{\'a}$^{1}$\\
$^{1}$George Mason University $^{2}$Drake University\\
{\tt\small \{yli44, nrajabi, sshres2, kosecka\}@gmu.edu, md.reza@drake.edu}
}

\maketitle

\begin{abstract}
The image annotation stage is a critical and often the most time-consuming part required for training and evaluating object detection and semantic segmentation models. 
Deployment of the existing models in novel environments often requires detecting novel semantic classes
not present in the training data. Furthermore, indoor scenes contain significant viewpoint variations, which need to be handled properly by trained perception models. 
We propose to leverage the recent advancements in state-of-the-art models for bottom-up segmentation (SAM), object detection (Detic), and semantic segmentation (MaskFormer), all trained on large-scale datasets. 
We aim to develop a cost-effective labeling approach to obtain pseudo-labels for semantic segmentation and object instance detection in indoor environments, with the ultimate goal of facilitating the training of lightweight models for various downstream tasks. 
We also propose a multi-view labeling fusion stage, which considers the setting where multiple views of the scenes are available and can be used to identify and rectify single-view inconsistencies. 
We demonstrate the effectiveness of the proposed approach on the Active Vision dataset~\cite{Ammirato2017ADF} and the ADE20K dataset~\cite{Zhou2017ScenePT}. 
We evaluate the quality of our labeling process by comparing it with human annotations. 
Also, we demonstrate the effectiveness of the obtained labels in downstream tasks such as object goal navigation and part discovery. 
In the context of object goal navigation, we depict enhanced performance using this fusion approach compared to a zero-shot baseline that utilizes large monolithic vision-language pre-trained models.  

\end{abstract}

\section{Introduction}

\begin{figure}[ht!]
\centering
\includegraphics[width=0.45\textwidth]{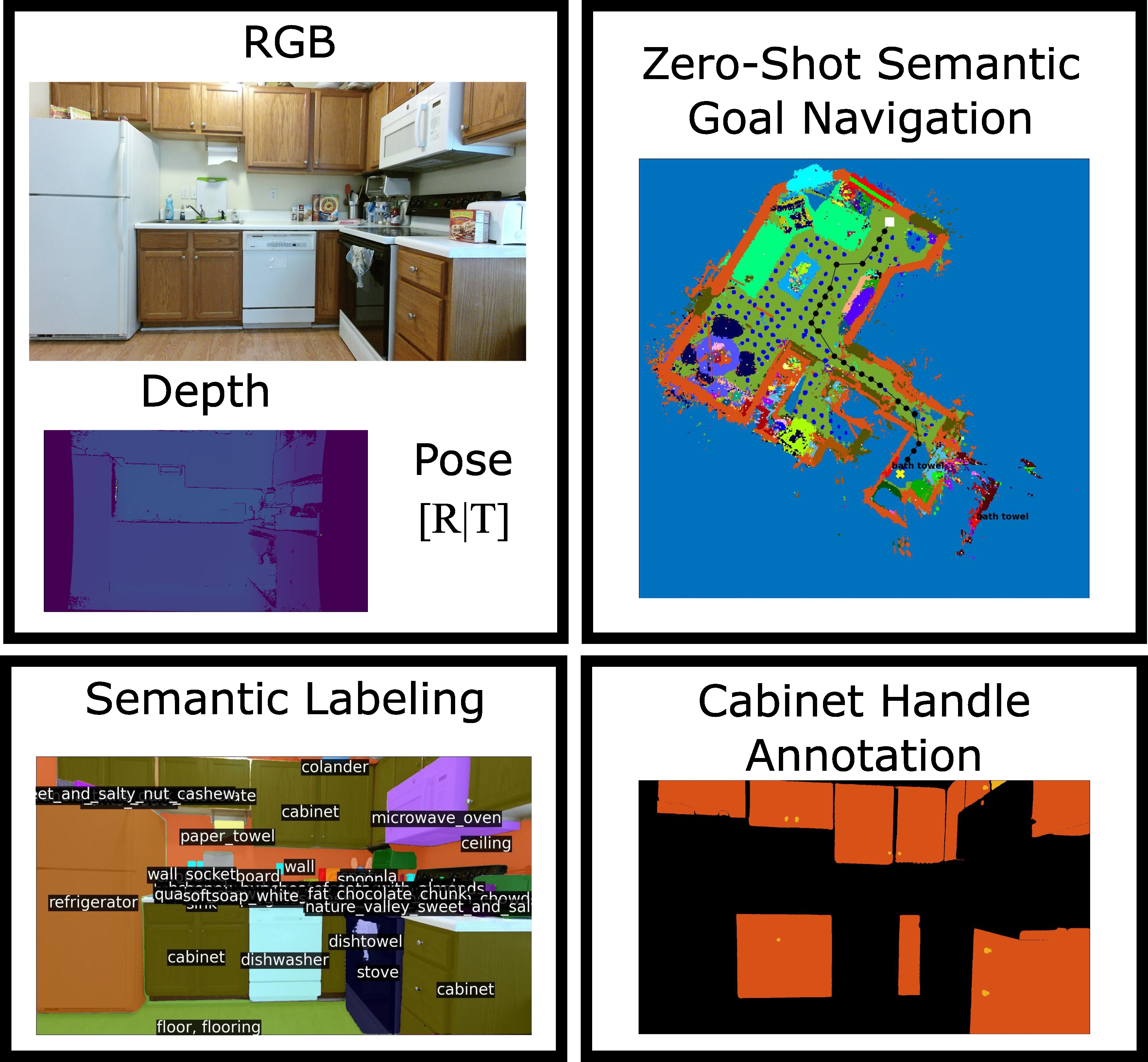}
\caption{
\textbf{This diagram gives an overview of our contributions.}
Starting from an RGB-D dataset, we propose a labeling approach for semantic segmentation annotations.
On top of the semantic segmentation results, we additionally proposed two downstream tasks for robot navigation.
We build top-down-view semantic maps and use them for zero-shot semantic-goal navigation.
We proposed an object part segmentation task for the 'cabinet handle' related to the robot mobile manipulation task. 
}
\label{fig:title}
%\vspace{-.5em}
\end{figure}

In recent years, there has been an increased interest in the design and evaluation of embodied agents (e.g., mobile robots) that operate in indoor environments such as restaurants, households, and hospitals. These embodied agents often need to perform tasks such as object goal navigation or fetch and delivery tasks, both of which require navigating to objects specific by natural language. Towards this end, various indoor scene datasets of real-world interiors have been recently used \cite{replica19arxiv, Ammirato2017ADF, Yadav2022HabitatMatterport3S}. 
% As we seek to develop intelligent household robots that can help us with our daily activities, good datasets of 3D models are becoming more important.
The perception in these tasks relies on the ability of the agent to detect objects of interest. 
% \deleted[id=SS]{Such to train AI agents to navigate to object goals 
% specified by natural language and to follow natural language instructions. }
% Recent work on acquiring and semantically annotating datasets of real-world spaces has significantly accelerated research into embodied AI agents that can perceive and navigate  within realistic indoor scenes. 
Obtaining dense semantic annotation of indoor scenes is challenging and time-consuming; 
The tasks of instance segmentation and semantic segmentation require pixel-level accuracy.
For example, it has been reported that labeling HM3D dataset~\cite{Yadav2022HabitatMatterport3S} took over 14,200 hours of human effort for annotation and verification by 20+ annotators. Since the indoor environments have a more diverse set of objects compared to the autonomous driving datasets, choosing the object category from over 1K object categories adds to the complexity of the task. 

% \deleted[id=JK]{Many previous datasets rely on utoronto-developed auto-labeling tools to accelerate labeling process so that only a course bounding box is needed, accurate mask around is not required.
% But this still requires human effort to recognize the objects.
% For a image taken from a broad view, it usually contains more than 50 small objects and takes more than 15 minutes of human labeling.
% Also, these auto-labeling tools cannot handle cases when the object are separated into small parts because of its appearance or occlusion.
% People have tried to % cite HM3D-Semantic, 3D scene graph paper
% But this requires expensive 3D cameras and there is a lot of errors when the 3D model has errors for objects like clothes or mirrors.
% %
% % start talk about SAM and Detic
% % thanks to the recent progress of open-set recognition and vision foundation models
% Recently, Meta AI proposed SAM, a foundational model for image segmentation.
% It takes in various prompt, points, bounding boxes and give out a mask segmentation for the object of interest.
% Besides, a powerful open-set object detector Detic is able to detect 21K object categories robustly.
% }

% start to talk multiview Reza's propagation paper

In this paper, we explore the effectiveness of fusing the predictions from existing state-of-the-art models 
% an auto-labelling approach for images of indoor scenes. For our approach, we propose to use the state-of-the-art models 
for bottom-up class-agnostic image segmentation~\cite{SAM_Kirillov2023SegmentA}, object detection~\cite{Detic_Zhou2022DetectingTC}  and
semantic segmentation~\cite{Maskformer_Cheng2021PerPixelCI} to obtain instance and semantic segmentation. While the class agnostic image segmentation model SAM~\cite{SAM_Kirillov2023SegmentA} produces accurate boundaries, 
it requires either a good bounding box hypothesis for the classes or a dense point initialization of the input image to generate segmentations. Hence, we exploit the accurate boundaries of SAM, jointly with the object detection model~\cite{Detic_Zhou2022DetectingTC} and the semantic segmentation model~\cite{Maskformer_Cheng2021PerPixelCI} to get labels and masks for foreground and background classes respectively. We demonstrate the effectiveness of this labeling approach by evaluating the resulting labels using human-annotated labels and through the performance of agents that utilize our approach for various downstream tasks. Our contributions are as follows:
\begin{itemize}
    \item We design a labeling approach that fuses predictions from the state-of-the-art semantic segmentation and object detection models to obtain semantic labels for the class-agnostic image segmentation. 
    \item We further refine the single-view labeling by enforcing multi-view semantic consistency on Active Vision Dataset~\cite{Ammirato2017ADF} (densely sampled multi-view dataset of indoor scenes) and evaluate the effectiveness of our approach by directly comparing it to the labels obtained with human annotations.\footnote{The resulting annotated dataset will be made publicly available.}
    \item We demonstrate the effectiveness of our approach in evaluating the performance of the object goal navigation task and part discovery on the AVD dataset, which utilizes the fused segmentation results. 
\end{itemize}

\section{Related Work}

% \deleted[id=SS]{We review several recent as well as more traditional approaches for endowing images with semantic information. For the task of semantic segmentation, traditional supervised learning method proceed use available datasets.} 

\noindent \textbf{Label Propagation.}
The idea of propagating labels between different views of the environment has been used extensively as a means of obtaining additional 
training examples for either contrastive learning or fine-tuning existing models semantic segmentation models. 
Label propagation has been used for scene completion in indoor scenes \cite{Zhi2021InPlaceSL, Reza2019AutomaticAF} and in videos \cite{Badrinarayanan_CVPR_2010, Reza2019AutomaticAF}. 
% \cite{Reza2019AutomaticAF} propose a fully automatic method for 
labeling pixels covering a range of categories from small objects to large furniture and backgrounds. 
Authors in ~\cite{Zhi2021InPlaceSL} model the 3D environment implicitly as a neural radiance field. 
Label propagation has also been used for gathering more training data either by directly using the available motion information ~\cite{Frakiadaki_BMVC21} or by training models to predict motion~\cite{Zhu_2019_CVPR}. However, the approaches that utilize label propagation in one form or another either assume that at least a few human-annotated labeled frames are available, keep the number of labels fixed and relatively small, and strive to train or fine-tune a single model. 
% Obtaining human annotation can be a very cumbersome task while utilizing only a single perception model restricts us from gaining knowledge about classes that may exist in other models. 

\noindent \textbf{Domain Adaptation.} Additional labels have been obtained in the past by training a model in a source domain of a particular environment (e.g., indoor) and then adapting the trained model to another domain of the same environment~\cite{pmlr-v80-hoffman18a}. While this does increase the performance of the model for the known classes, the classes novel to the model are missed entirely. 
Most, if not all, unsupervised or self-supervised domain adaptation approaches to target the autonomous driving domain, which has smaller view-point variations and less challenging occlusions~\cite{Yang-wacv21,Tranheden-wacv21,Zhang-cvpr21} and does not utilize association between multiple views from the environment.

% {\bf JK Some references from CLIP NAV, Clip on Wheels etc how these modeks are used for navigation}
% \noindent \textbf{open vocabulary object detection}
% Talk about CLIP, LSeg and Detic. 
% {\bf JK - this is ok, but I would not emphasise the language part as
% we are not doing any vision langauge navigation. Do not make is a section}
% \noindent \textbf{Vision Language Navigation}\\

\noindent \textbf{Open-Vocabulary Object Detection.} 
The powerful large multi-modal vision and language models~\cite{clip} have been used effectively for semantic segmentation~\cite{lseg} or as open vocabulary object detectors in zero-shot setting in multiple recent works on object goal naviation~\cite{clip-nav, clip-on-wheels, zson, vlmaps}. 
%  subsequent to the CLIP~\cite{clip} introduction, for the purpose of zero-shot navigation. 
These approaches mainly leverage (1) open-vocabulary (as opposed to the traditional fixed-category methods), (2) joint embedding space (for vision and language representations), and (3) large-scale pre-training on 400 million images of models like CLIP. For instance, CLIP-Nav~\cite{clip-nav} has applied this technique to a variation of Vision-and-Language Navigation (VLN) task, 
% by using CLIP for contrastive view selection by encoding 4 non-overlapping RGB views (each representing a unique navigable direction to go next) using CLIP \textit{vision}-encoder separately, and encoding the current phrase/segment of the navigation instruction (to be executed/grounded) using CLIP \textit{language}-encoder. Then, computing the dot-products between the encoded phrase and all 4 encoded views and select the view with the highest score as the best match in the joint semantic space. 
while CLIP on Wheels (CoW) ~\cite{clip-on-wheels} approach introduces a model for zero-shot semantic goal navigation. It combines classical frontier exploration with the generation of CLIP heatmaps, achieved by calculating the dot product between egocentric visual embeddings and target object text embeddings. Upon detecting the target object during exploration, the approach projects its pixels onto a map, facilitating straightforward robot navigation toward the goal.

While zero-shot navigation-based approaches have been aiming to eliminate the need for domain adaptation and enable open-vocabulary navigation, there is still significant room for improvement of observed poor zero-shot performances. We think a fundamental issue is the reliance on dot-product scores when comparing the holistic vision and language-embedded representations in joint high-dimensional embedding space. 
% Another issue with models like CLIP is the proprietary corpus/datasets used for their pre-training. 
Due to the nature of the representations these models learn, it is difficult to scrutinize the details of concepts/categories covered during pre-training and the frequency of those samples. We hypothesize that fusing the predictions from the state-of-the-art perception models and building a richer semantic map, including long-tail object instance predictions, could be more effective in downstream tasks as a trade-off between the fixed-category and zero-shot open-vocabulary approaches.

% In computer vision, existing works that build instruction-following agents can be broadly categorized into two classes: End-to-end trained models, which are separately explored in each specific research topic. 
% For example, the vision-language navigation task and Habitat require the embodied AI agent to follow natural language instructions and take a sequence of actions to complete goals in visual environments.

\section{Preliminaries}

We briefly review three main components of our labeling approach. 

\noindent \textbf{Segment Anything Model (SAM)}~\cite{SAM_Kirillov2023SegmentA} 
SAM is a Vision Transformer (ViT) based model trained extensively on a very large dataset of 1 billion masks from their SA-1B dataset. 
SAM accommodates diverse input prompts, including points, bounding boxes, and dense point grids, and returns boundaries corresponding to the
prompts. Such prompts provide a coarse estimation of what to segment in an image, which enables SAM to undertake a wide range of segmentation tasks without the need for additional training. A notable strength of SAM is its innate grasp of what constitutes an object, which enables zero-shot generalization to 
unfamiliar objects that are commonplace in indoor environments.

\noindent \textbf{Object Detector (Detic)}~\cite{Detic_Zhou2022DetectingTC} 
The Detic model is an object detector that can efficiently identify 21,000 object classes, including many previously considered challenging. 
Leveraging the idea of training object detectors in a Weakly-Supervised manner, Detic is trained on the ImageNet-21K dataset and sidesteps the traditional reliance on assigning labels to bounding boxes. 
Using text embeddings from CLIP~\cite{clip} and a classifier based on concept recognition, Detic recognizes objects beyond predefined categories, enhancing its adaptability. The model accommodates various categories from various datasets such as COCO~\cite{Lin2014MicrosoftCC}, OpenImages~\cite{Shao2019Objects365AL}, and LVIS~\cite{Gupta2019LVISAD}. Because of its use of CLIP text embeddings for class specification, the model can also be adapted to incorporate a custom vocabulary.

\noindent \textbf{MaskFormer}~\cite{Maskformer_Cheng2021PerPixelCI} 
The MaskFormer model presents a novel approach to tackling semantic and instance segmentation tasks using a unified mask classification framework. 
MaskFormer employs a Transformer decoder to compute a set of pairs comprising class predictions and mask embeddings, with the mask embeddings used to compute binary mask predictions.
The model excels in handling datasets with diverse categories. 
It achieves a new state-of-the-art performance on the ADE20K dataset, outperforming per-pixel models with similar backbones.

\begin{figure*}[ht!]
\centering
\includegraphics[width=1.0\textwidth]{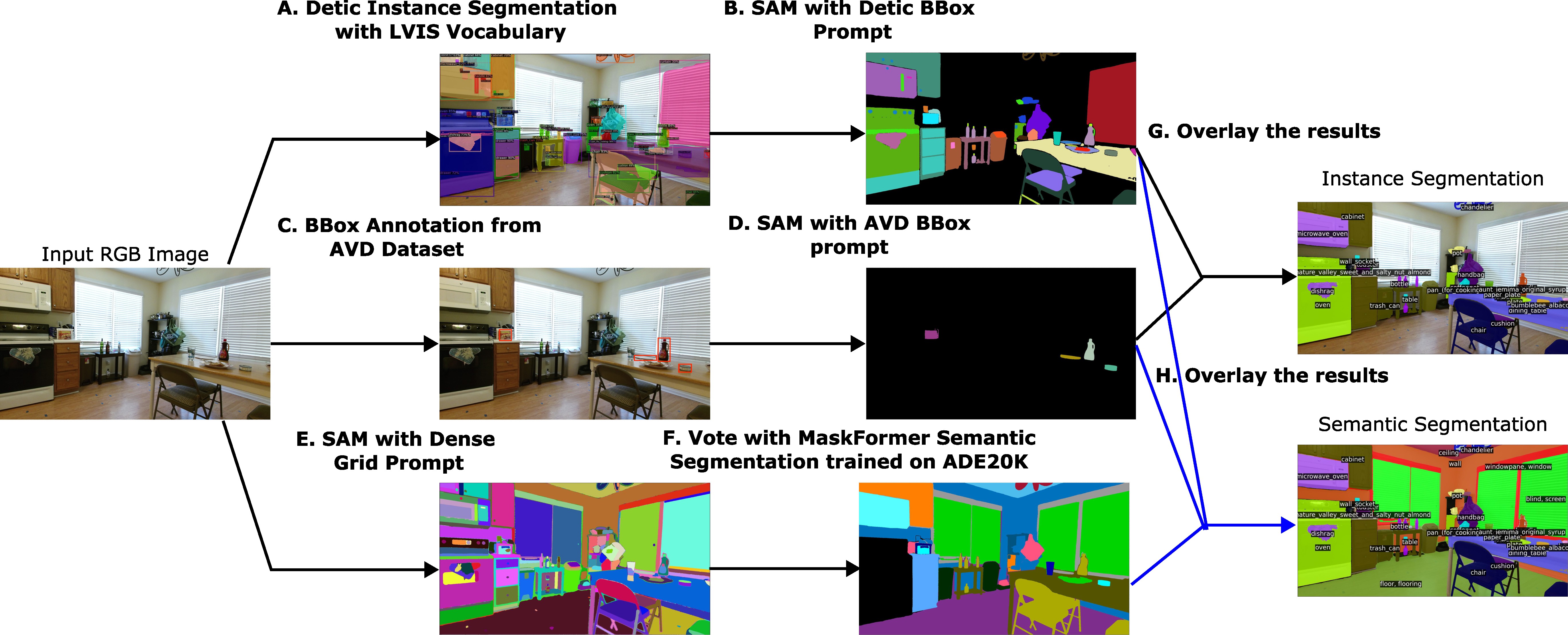}
\caption{
\textbf{This diagram gives an overview of our labeling approach at the single-view labeling stage.}
Given an input image, we generate masks for foreground classes utilizing Detic and SAM at steps (A) and (B).
If any classes with manual bounding boxes are available, we generate masks utilizing SAM at steps (C) and (D).
We generate masks for the entire image utilizing MaskFormer and SAM at steps (E) and (F).
We overlay the results of foreground class masks on top of the entire image's mask and achieve the semantic segmentation and instance segmentation annotations.
}
\label{fig:approach}
%\vspace{-.5em}
\end{figure*}

\section{Labeling Approach}
The proposed labeling approach is tailored for indoor scene datasets featuring high-resolution images of cluttered environments, including diverse object types. 
% The approach can be divided into single-view labeling and multi-view verification. 
The single-view labeling stage leverages the open-vocabulary object detection Detic~\cite{Detic_Zhou2022DetectingTC}, the state-of-the-art semantic segmentation model MaskFormer~\cite{Maskformer_Cheng2021PerPixelCI}, and the foundational segment anything model SAM~\cite{SAM_Kirillov2023SegmentA} to annotate all the images with pixel-wise labels.
When presented with depth images and corresponding camera poses, the integration of the multi-view verification stage can further enhance the results through the merging of single-view stage results.
The stages are explained in detail below.

\subsection{Single-view Labeling Stage} \label{sec:single_view}

The initial stage of our labeling pipeline involves labeling single views from the environment independent of the other views. 
This stage encompasses three distinct branches that can be processed separately: 
(i) Generate masks for the entire image utilizing the semantic segmentation model (MaskFormer) and SAM; 
(ii) Generate masks for foreground classes utilizing object detection model (Detic) and SAM; 
(iii) If manual bounding boxes are available, generate corresponding masks utilizing SAM.

\begin{figure}[H]
\centering
% \vspace{-.5em}
\includegraphics[width=0.45\textwidth]{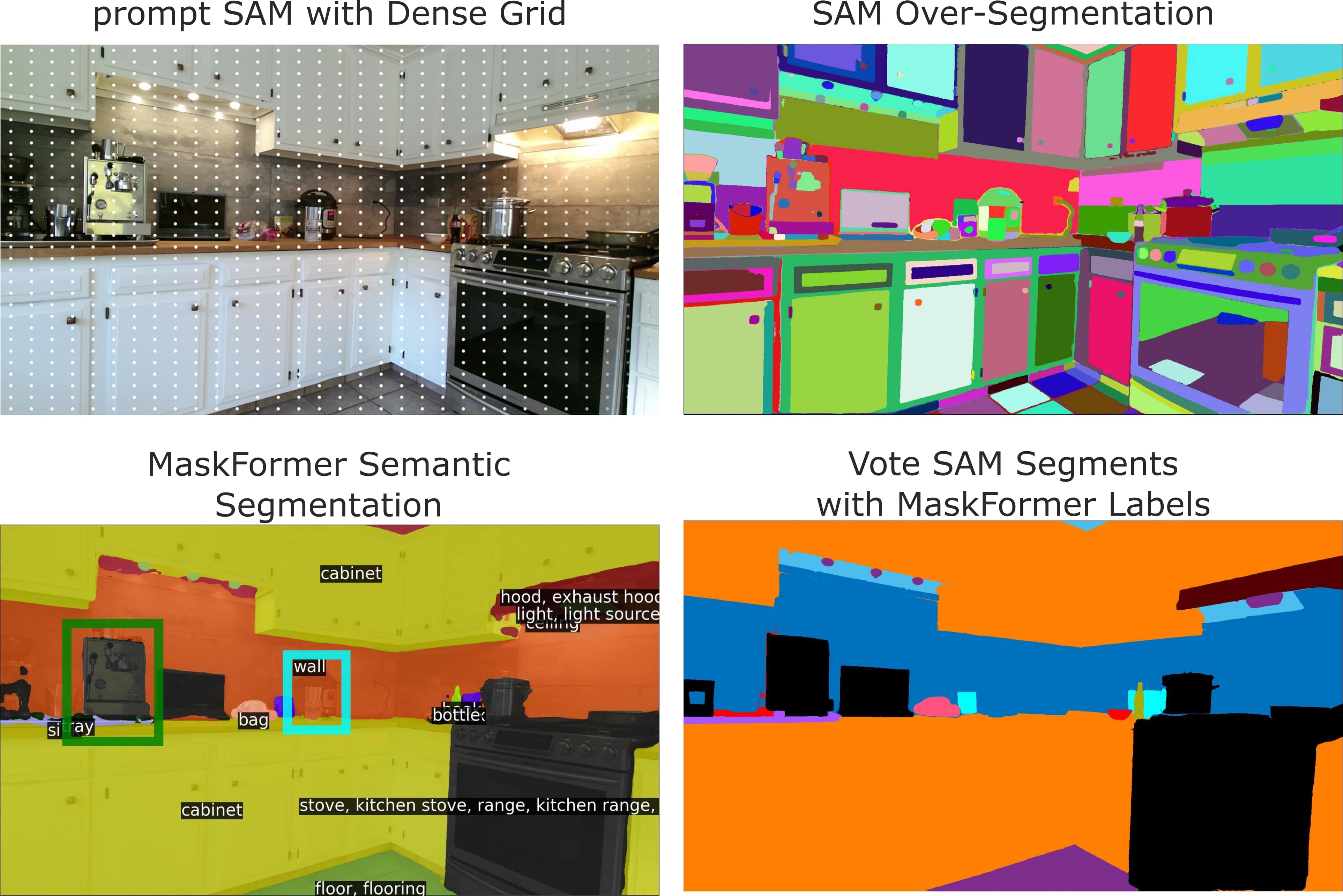}
\caption{
An example of labeling using Semantic Segmentation (MaskFormer) and SAM. MaskFormer produces good predictions for background classes but not so well for foreground classes: the coffee machine is misclassified as a `stove' (black bounding box), and the cooking pot (cyan bounding box) is missed entirely. 
}
\label{fig:maskformer_vis}
%\vspace{-.5em}
\end{figure}

\noindent \textbf{Semantic Segmentation with SAM}
We prompt SAM with dense point grids in the input image. This operation yields a collection of non-overlapping class agnostic segments. We run MaskFormer (trained on ADE20K~\cite{Maskformer_Cheng2021PerPixelCI}) to get the semantic segmentation of the image. We then assign labels to each segment generated by SAM by voting. Specifically, each segment gets votes per class based on the number of pixels that lie inside the segment and belong to the class based on MaskFormer results. The segment is assigned the class label with the highest vote. This step is illustrated in Fig.~\ref{fig:approach} (F). This step produces good results for background classes but misses a lot of foreground classes, as shown in Fig.~\ref{fig:maskformer_vis}. Hence, we opt to use an object detector for foreground classes.

\begin{figure}[h]
\centering
% \vspace{-.5em}
\includegraphics[width=0.45\textwidth]{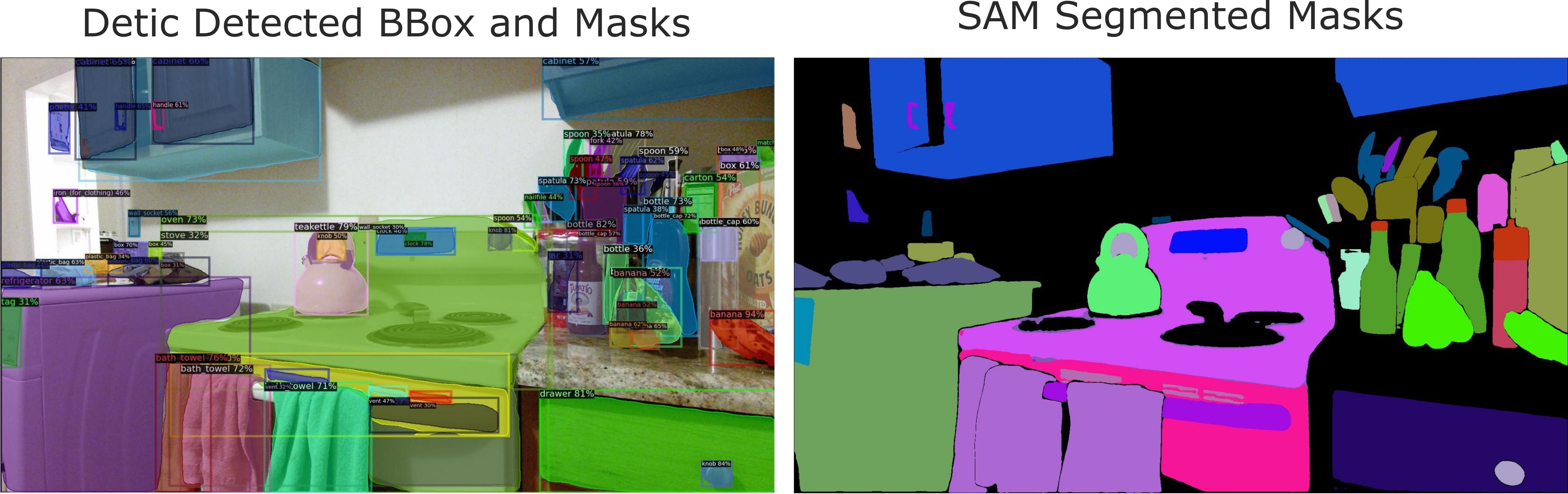}
\caption{
An example of labeling using Object Detector (Detic) and SAM. 
}
\label{fig:detic_vis}
%\vspace{-.5em}
\end{figure}

\noindent \textbf{Object Instance Detection and Segmentation with SAM}
Semantic segmentation models usually perform well on background classes but may lack the same accuracy for foreground classes. Furthermore, they may be trained with a limited vocabulary. For example, MaskFormer is trained on ADE20K, which lacks many common indoor objects like `a cooking pot' and `knife'.
To address these shortcomings, we employ Detic with LVIS vocabulary, which includes 1203 classes and covers many known indoor object classes.
However, the jointly predicted object bounding boxes and masks from Detic often do not have the desired level of accuracy, especially when compared with SAM results, as shown in Fig.\ref{fig:detic_vis}.
This is due to the fact that Detic is partially trained on ImageNet, a dataset without mask annotations. We utilize SAM to generate high-quality masks by using the bounding boxes as input prompts. 
In practice, we prompt SAM with bounding boxes and a point corresponding to the centroid of the masks from Detic, leading to high-quality object instance segmentation shown in Fig.~\ref{fig:detic_vis} and Fig.\ref{fig:approach} (B).
% {\bf If we can fit it would be good to have more examples of single bigger image depicting this. The results are quote impressive and constantly going to the big picture is not ideal. This way we can show results on more image, on images with larage clutter, etc} \\

% \noindent \textbf{Overlay Results}
%MaskFormer has other categories but we only use background. We assume that LVIS category covers all the objects other than the instances from AVD.
% Since some categories in the semantic and object detection branches exhibit overlap, we adopt a selective approach by assigning the semantic 

We select the background classes {\em floor}, {\em ceiling}, {\em door}, {\em blind}, and {\em wall} from the semantic segmentation model and discard other classes. The final result is obtained by superimposing foreground class masks, obtained from the use of object detection and SAM, on top of semantic segmentation results, achieving a comprehensive and refined final representation.

\subsection{Multiview Verification Stage}
The single-view annotation may produce incorrect predictions due to model prediction errors, especially in challenging scenarios such as partial occlusion of objects or irregular viewing angles. 
With the assumption that models perform well in most scenarios and that objects are less occluded, at least in some viewpoints, results from such views can rectify these prediction errors. Hence, we fuse masks from multiple views using a per-class voting approach.

Consider an image $I_k$ and its corresponding single-frame annotation $A_k$ that requires label verification. 
We designate two keyframes, namely $I_m$ and $I_n$, with their respective single-view annotations $A_m$ and $A_n$, respectively, which serve as reference frames for $I_k$. 
In practice, the selection of $I_m$ and $I_n$ is guided by their proximity to the viewpoint of $I_k$, thereby ensuring contextual relevance. 
It's worth noting that all frames apart from $I_k$ can be utilized as reference images in this process, provided they overlap with frame $I_k$.

We begin by projecting the annotations $A_m$ and $A_n$ into a 3D spatial context, utilizing the respective image poses $[R_m|T_m]$ and $[R_n|T_n]$ and their depth maps. The projections yield the corresponding 3D point clouds $P_m$ and $P_n$, respectively. Each point within $P_m$ and $P_n$ has an associated semantic label from $A_m$ and $A_n$. 
Using a 4-way connected component in $A_k$, we get the set of regions $\text{Reg}_k$ in $I_k$.
Subsequently, we traverse each region in $\text{Reg}_k$ within $A_k$, treating each region as a superpixel, similar to the approach outlined in~\cite{Reza2019AutomaticAF}. We then project $P_m$ and $P_n$ onto the $I_k$ frame using its  known camera pose $[R_k|T_k]$. 
To ensure the correctness of the projected points, we filter out erroneous projections by cross-checking the projected point's depth value with the depth of the original pixel.

% we compute the average of each region.

For each region in $\text{Reg}_k$, we calculate the votes for each class $c_j$ using projected pixels from the two reference point clouds $P_m$ and $P_n$. First, we calculate the number of pixels that were projected from $P_m$ to this region that has the class $c_j$. We normalize this value using the total number of pixels in the region to get the normalized score $f_m^{c_j}$ for class $c_j$. We perform similar steps to get score $f_n^{c_j}$ using the other reference point cloud $P_n$. The score $f_k^{c_j}$ is set to $1$ because the region is obtained using the class labels in $A_k$ itself. The final vote for class $c_j$ in the region is obtained by taking the average of the scores $f_m^{c_j}$, $f_n^{c_j}$ and $f_k^{c_j}$. The region is assigned the class which has the highest vote.

\begin{figure}[H]
\centering
% \vspace{-.5em}
\includegraphics[width=0.45\textwidth]{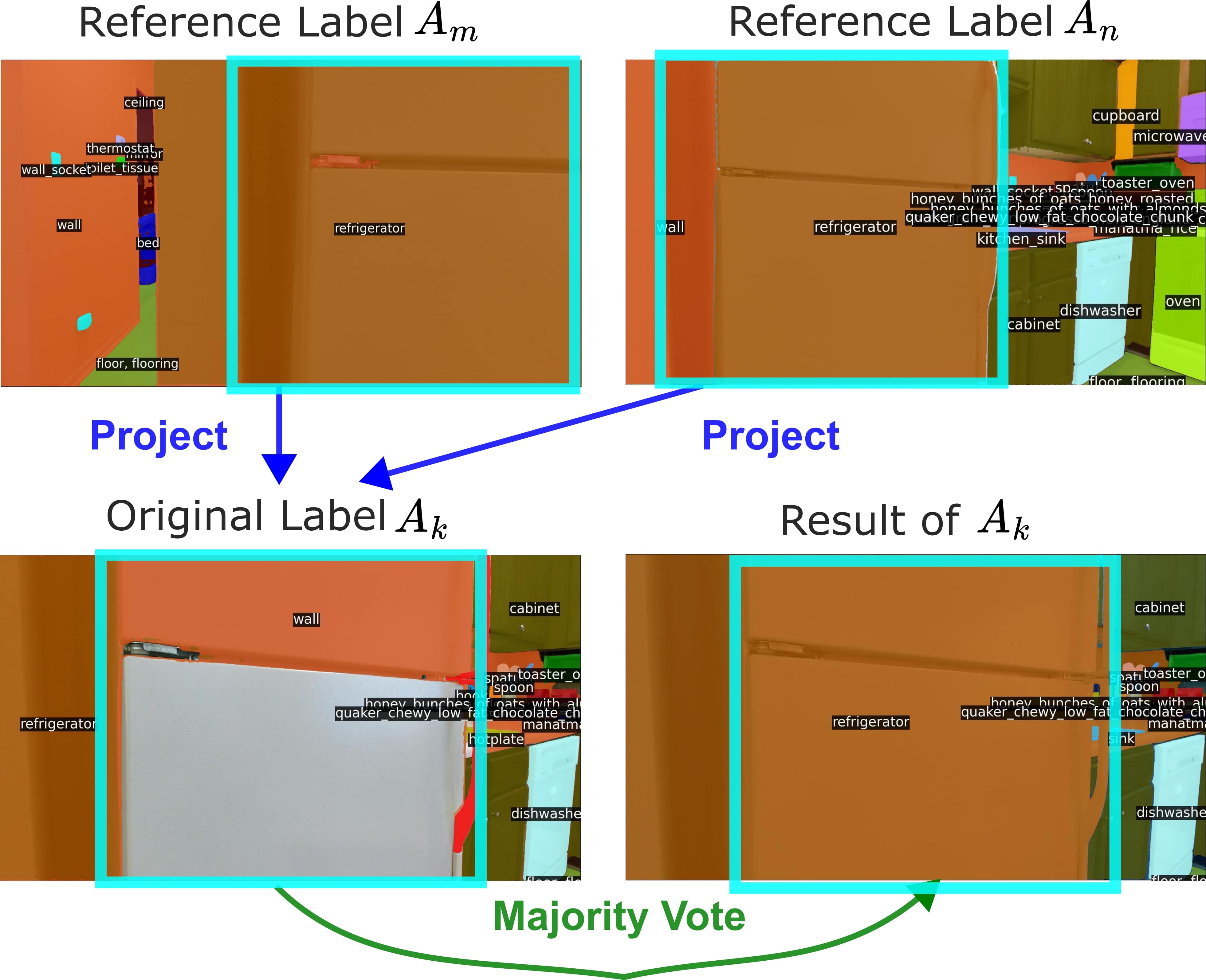}
\caption{
An example of Multiview Verification. The refrigerator (cyan bounding box) in view $A_k$ is originally labeled as wall and missing mask for some of its parts. The error is resolved by fusing labels from views $A_m$ and $A_n$ with the correct annotation for the class. 
}
\label{fig:mv_vis}
%\vspace{-.5em}
\end{figure}

Fig.~\ref{fig:mv_vis} illustrates the multiview verification process.

% If $\text{Reg}_k$ is assigned label $c_j$, it acquires a score of $f_k^{c_j}=1$; otherwise, it receives a score of $0$.
%To finalize the region's semantics, we take the average of the scores $f_k^{c_j}$, $f_m^{c_j}$, and $f_n^{c_j}$. 
% To finalize the region's semantics for class $c_j$, we take the average of the scores from all the reference views $f_k^{c_j}$, $f_m^{c_j}$, and $f_n^{c_j}$. 
% Subsequently, we perform an argmax computation across all categories to ascertain the semantic label for the given region.

\section{Downstream Tasks}

We next evaluate the performance of the obtained instance segmentation and semantic segmentation 
on two downstream tasks: part discovery and object goal navigation. 
% We introduce two  tasks besides the conventional semantic and instance segmentation tasks.
% We also present another task concerning the zero-shot semantic-goal navigation task, in which we demonstrate the effectiveness of our semantic map to be used in the relevant down-stream tasks 

\begin{figure*}[ht!]
\centering
\includegraphics[width=1.0\textwidth]{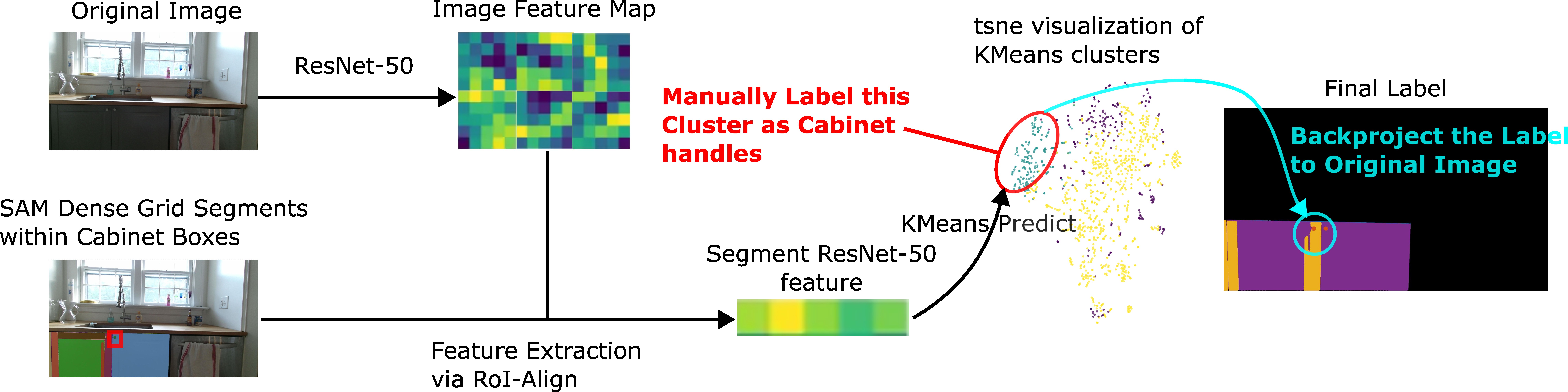}
\caption{
\textbf{This diagram gives an overview of the labeling cabinet handle.}
We choose SAM segments within the detected 'cabinet handle' bounding box.
Then, we extract ResNet50 features for these segments.
We cluster these feature points through KMeans and manually label the cluster containing 'cabinet handle' points.
We backproject the label to the original image to get the 'cabinet handle' annotations.
}
\label{fig:exp_2_approach}
\end{figure*}

% Currently, the VLN task is primarily defined on the Matterport3D dataset, where the fine-grained category labels are absent, and the referenced objects in the instructions are not aligned with the object labels.
% To address these limitations, we devise an alternative approach within our dataset. 
% This involves generating a set of 10 instructions for each scene, considering the objects present within the scene. 
% This new task bridges the gap between instructions and scene objects, offering a more coherent and contextually aligned environment for evaluating VLN approaches.

\subsection{Object Part Discovery}
\label{sec:obj_part_seg}
Part discovery is important for mobile manipulation tasks such as opening doors, cabinets, bottles, or adjusting thermostats. Since the shape of hinges and handles can differ significantly across different houses,
we describe a simple approach for detecting these parts in a particular environment. 
% especially in cases these cannot be detected by the state-of-the-art object detector. 
% We consider scenarios where a robot undertakes intricate actions such as uncapping a bottle, operating cabinet handles, and adjusting thermostat controls to modify temperature settings.
% To facilitate the evaluation of detecting the fine-grained parts of common objects, 
% we provide labels for 'bottle caps' and 'cabinet handles'. 
% These annotations offer a means to gauge the robot's prowess in nuanced articulation and interaction with the environment.
We demonstrate this approach to the detection of cabinet handles. 
% Obtaining the 'bottle cap' annotation is relatively straightforward, as it is included within the LVIS categories.
% Leveraging Detic's adeptness in bottle cap detection, we prompt SAM with the detected bounding boxes and generate masks for 'bottle cap'.
% Annotating cabinet handles is more challenging given that 'cabinet handle' is not covered by existing dataset classes. 
We use SAM to generate an over-segmentation of the image, followed by cabinet detection.
SAM's segmentation typically separates the cabinet into three distinct regions: the handle, the drawer, and the cabinet's main body. Focusing on the segments within the predicted 'cabinet' bounding boxes from Detic,
We first extract the ResNet-50 feature map of the entire image and extract the feature vector for each cabinet 
segment using RoIAlign. This is followed by k-means clustering with these features and 
manually identifying the cluster containing points corresponding to cabinet handles.
We project data points within this cluster back to the images, thereby identifying segments of cabinet handles.
In practice, we apply k-means clustering to each scene individually.
We use 3 to 5 clusters considering the cabinet's appearance in the environment. 

Fig.~\ref{fig:exp_2_approach} illustrates the object part discovery approach.

\subsection{Object Goal Navigation}
Next, we evaluate the effectiveness of obtained semantic labels for the semantic object goal navigation task. 
More specifically, we focus on long-tail object instances and compare them to the state-of-the-art 
navigation approaches that use CLIP joint embedding space of large multimodal vision-language models~\cite{clip} to localize objects in images and maps like VLMaps~\cite{vlmaps,clip-nav,clip-on-wheels}.

% build a map
We first build a top-down semantic map for each scene of AVD using the semantic segmentation labels.
The semantic map is a metric map of size $m \times m$ where each cell's value is in the range $[0, N]$.
Each value from 1 to $N$ corresponds to one of the N object categories, and 0 refers to the undetected class. 
Each cell is a $5cm \times 5cm$ region in the real world.
Given an RGB-D view, we build the semantic map by projecting semantic segmentation images to a 3D point cloud using the available depth maps, and the robot poses, discretizing the point cloud into a voxel grid and taking the top-down view of the voxel grid.
The semantic map depends on the majority category of the points located at the top grid of each cell.
We summarize the semantic categories that exist in the scene by going through each semantic category and localizing the corresponding cells on the map.

% generate instructions
For each AVD scene, we collect a semantic goal to evaluate object navigation. Then, we randomly specify a starting position of the robot in the scene and then pick a target object to navigate to.
% among 30 object categories as subgoal object types. 
% The robot is required to navigate to these four subgoals sequentially. 
% When the robot reaches one subgoal category in each sequence of subgoals, it should call the stop action to indicate its progress. 
We consider the navigation to the target successful when the agent reaches the nearest navigable pose of the target object. (see Section~\ref{sect:exp_obj_nav} for experiment details)
% To evaluate the long-horizon navigation capabilities of the agents, we compute the success rate (SR) of continuously reaching one to four subgoals in a sequence.

\section{Experiments}

% \begin{table*}[t]
% \caption{mIoU($\%$) of semantic segmentation methods (higher is better)}
% \label{tab:sseg}
% \centering
% \begin{tabular}{l|c|c|c|c}
% Dataset & \multicolumn{2}{c}{ADE20K-indoor} & \multicolumn{2}{c}{AVD-GT} \\
% \hline
% Method & mIoU ($\%$) & mIoU-small ($\%$) & mIoU ($\%$) & mIoU-small ($\%$) \\
% \hline
% MaskFormer~\cite{Maskformer_Cheng2021PerPixelCI} & 55.7 & 62.3 & 58.6 & 60.2 \\
% MaskFormer+SAM~\cite{SAM_Kirillov2023SegmentA} & 56.8 & 63.0 & 64.9 & 70.7 \\
% MaskFormer+SAM+Detic (Ours) & \textbf{58.3} & \textbf{63.6} & 64.2 & 92.8 \\
% MaskFormer+SAM+Detic+MV (Ours) & -- & -- & \textbf{65.4} & \textbf{97.7} \\
% \end{tabular}
% \end{table*}

\noindent \textbf{Datasets.} 
We use the Active Vision Dataset (AVD)~\cite{Ammirato2017ADF} and the ADE20K~\cite{Zhou2017ScenePT} dataset for our experiments. We demonstrate and evaluate our labeling approach on AVD and ADE20K, while we only use AVD for downstream tasks.

AVD is comprised of 20 distinct environments that cover a variety of realistic indoor scenes, ranging from kitchens and offices to living rooms and bedrooms. We use the AVD dataset as a basis for our labeling effort because it captures real-world indoor scenarios characterized by intricate clutter and the presence of open-set objects, posing significant challenges to existing perception models. 
AVD also contains realistic depth images captured by Kinect V2, and the poses associated with the observations are available, making it an ideal dataset for downstream robotics applications. 
Furthermore, AVD has high-resolution RGB images, making annotation of fine-grained objects more feasible. The semantic categories of AVD are consistent with  ADE20K~\cite{Zhou2017ScenePT} and LVIS~\cite{Gupta2019LVISAD} datasets. To evaluate the labeling approach, we manually labeled the first ten images of each scene with the instance and semantic segmentation. 
% by going through each region of the auto-labeling result and correcting the category if the original annotation was wrong.
In total, we labeled 200 images for 20 scenes. We name this annotated subset of the dataset AVD-GT.

% JK This distinction is notable for its adept handling of challenges posed by mirrors, cluttered clothes, and rendering artifacts.

% Furthermore, we extend the bounding box annotations of BigBIRD objects in AVD to masks, opening avenues to potential robotics applications, such as instance recognition. 

ADE20K has room-type labels for each image. Since the proposed approaches are tailored for indoor scenes, owing to the classes selected, we only select images labeled as {\em bathroom}, {\em bedroom}, {\em kitchen}, {\em living room}, {\em office}, {\em dining room}, {\em hotel room}, {\em dorm room}, {\em home office} or {\em waiting room}. We used 440 indoor scene images from the ADE20K for the evaluation. We name this subset of the dataset ADE20K-indoor.

\begin{table*}[t]
\caption{mIoU($\%$) of semantic segmentation methods (higher is better)}
\label{tab:sseg}
\centering
\begin{tabular}{l|c|c|c|c}
Dataset & \multicolumn{2}{c}{ADE20K-indoor} & \multicolumn{2}{c}{AVD-GT} \\
\hline
Method & mIoU ($\%$) & mIoU-small ($\%$) & mIoU ($\%$) & mIoU-small ($\%$) \\
\hline
MaskFormer~\cite{Maskformer_Cheng2021PerPixelCI} & 55.7 & 62.3 & 58.6 & 60.2 \\
MaskFormer+SAM~\cite{SAM_Kirillov2023SegmentA} & 56.8 & 63.0 & 64.9 & 70.7 \\
MaskFormer+SAM+Detic (Ours) & \textbf{58.3} & \textbf{63.6} & 64.2 & 92.8 \\
MaskFormer+SAM+Detic+MV (Ours) & -- & -- & \textbf{65.4} & \textbf{97.7} \\
\end{tabular}
\end{table*}

\noindent \textbf{Implementation Details.}
For the single-view annotation process described in Section~\ref{sec:single_view}, we set the parameters of the SAM model, namely the mask IoU threshold and the mask prediction stability, to 0.86 and 0.92, respectively. 
Considering AVD's high-resolution RGB images and the abundance of small objects within each frame, we prompt SAM with a densely spaced point grid comprising $64 \times 64$ in the input image. 
The complete labeling is carried out on the entire AVD dataset and takes approximately 120 hours, utilizing two A100 GPUs. 

\subsection{Labeling}

We compare the labeling results with human-labeled ground truths. Originally, the AVD dataset has approximately 20,000 RGB-D images and a total of approximately 50,000 bounding boxes. We provide semantic segmentation annotations for all images covering more than 300 categories.
Distinguishing itself from the HM3DSem dataset~\cite{Yadav2022HabitatMatterport3S}, where 3D mesh models undergo time-consuming semantic labeling and where semantic segmentation images are rendered at each viewpoint, our pipeline delivers high-quality single-view annotations with lesser computational requirements.

\noindent \textbf{Metrics.}
We use \textit{mIoU} as the metric for evaluating the segmentation results, which is the average of the intersection-over-union (IoU) of all the categories.
% the ratio of the overlapping area between the predicted segmentation mask and the ground truth mask to the total area covered by both masks, {\bf mIoU} then averages the IoU values across all categories. 
We also compute \textit{mIoU-small}, which is the mIoU of only the small object categories.
For ADE20K-indoor, we choose {\em bottle}, {\em plant}, {\em lamp}, {\em glass}, {\em flower}, and {\em vase} as the small object categories. For AVD, we add {\em knife}, {\em bowl}, {\em plate}, and {\em wall socket} into the small object category list. We compare the following approaches:\\
\noindent\textbf{MaskFormer}: A semantic segmentation model trained on ADE20K. We use the publicly available code and the trained weights of this model~\cite{Maskformer_Cheng2021PerPixelCI}.\\
\noindent\textbf{MaskFormer+SAM}: Prompt SAM~\cite{SAM_Kirillov2023SegmentA} with a dense grid to get an over-segmentation of the input image. The assignment of labels to each segment is determined by voting as described in Sec.~\ref{sec:single_view}.\\
\noindent\textbf{MaskFormer+SAM+Detic (Ours)}: On the outcome of the 'Mask+SAM' approach, we overlay the results of Detic detections.\\
\noindent\textbf{MaskFormer+SAM+Detic+MV (Ours)}: On the outcome of the 'Mask+SAM+Detic' approach, we further apply multiview verification if depth images and camera poses are available.

Note that ADE20K-indoor does not provide depth images and camera poses. Hence, we did not evaluate the MaskFormer+SAM+Detic+MV approach with it. 
Besides, as the ADE20K semantic category set is a subset of LVIS categories, we create new annotations for MaskFormer and MaskFormer+SAM approaches when evaluating them with AVD.
The new annotations are created by finding the synonyms of the LVIS class in the ADE20K category set.
If no synonyms are found, e.g. {\em knife}, we label it as class {\em void}. We show these results in Table~\ref{tab:sseg}. The SAM-based approach consistently improves the model's performance by having better boundary segmentation of the segments. MaskFormer+SAM outperforms MaskFormer+SAM+Detic on the AVD dataset.
This is because the labels of AVD-GT cover fewer classes with the ADE20K vocabulary. 
%Fig.~\ref{fig:exp1_ade20k} shows a semantic segmentation result on an image from the ADE20K dataset.

\subsection{Object Part Discovery}
\begin{figure}[ht!]
\centering
\includegraphics[width=0.48\textwidth]{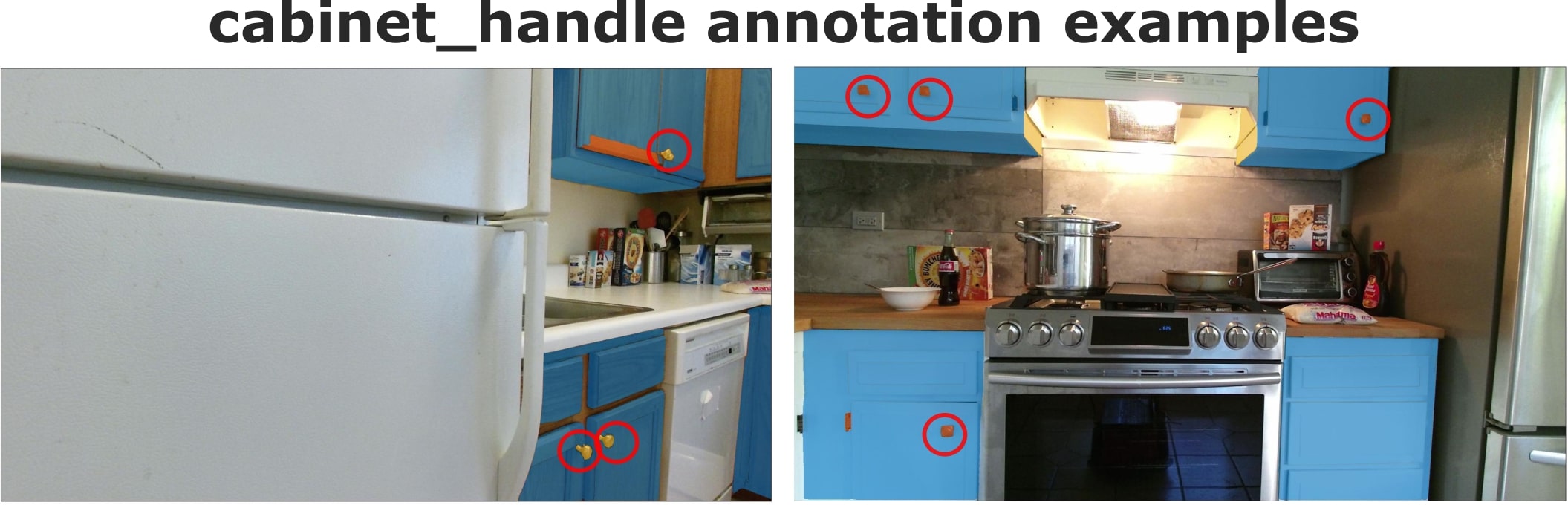}
\caption{
This figure gives examples of cabinet handles marked as red circles discovered through our clustering process.
}
\label{fig:vis_exp_2_cap_and_handles}
\end{figure}

% We create the 'bottle cap' dataset with 3000+ images and 6000+ bounding boxes and associated masks.
We create the 'cabinet handle' dataset with 1000+ images and 3000+ bounding boxes and associated masks.
Fig.~\ref{fig:vis_exp_2_cap_and_handles} shows the successful annotation examples.
% Detic failed to detect a few bottle caps.
We suppose the feature clustering approach applied for the 'cabinet handle' can be easily utilized for bottle part segmentation.

\begin{figure*}[ht!]
\centering
\includegraphics[width=0.95\textwidth]{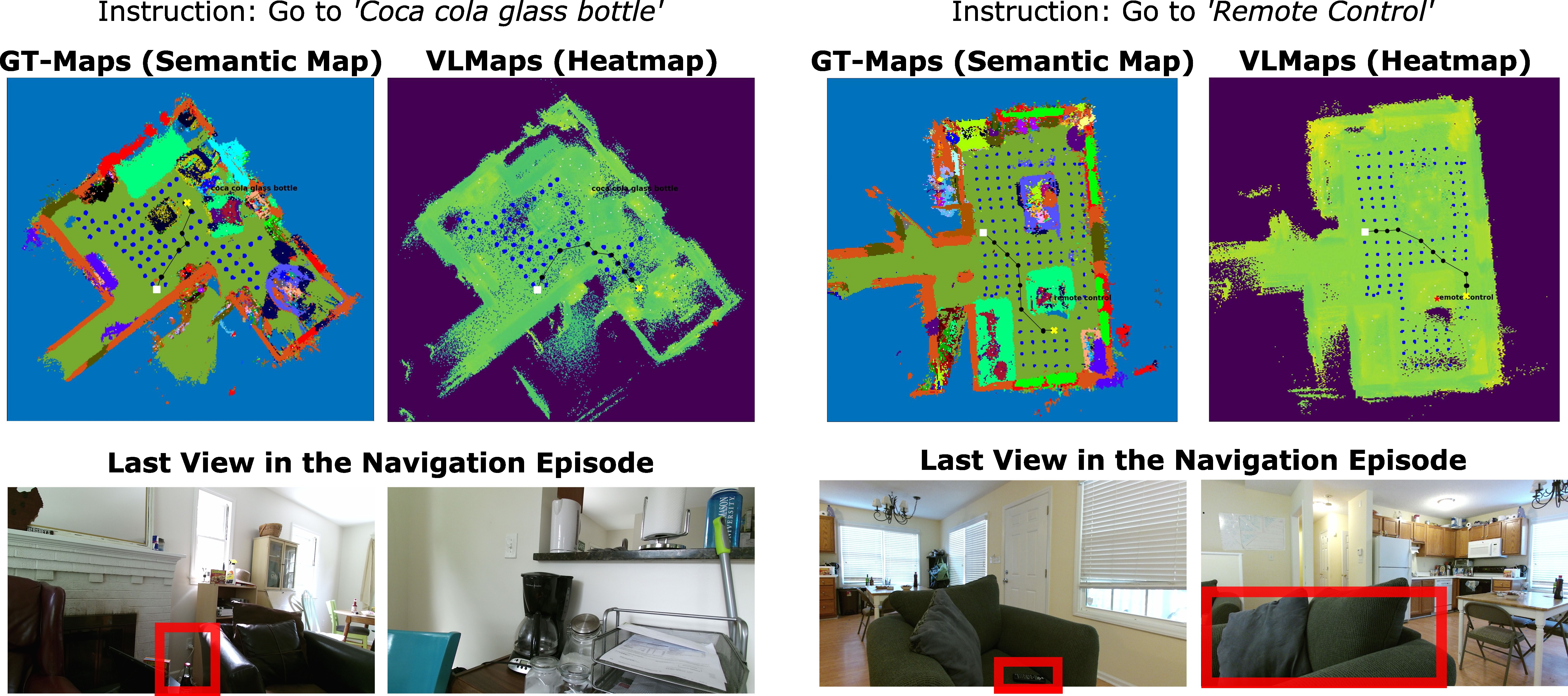}
\caption{
Qualitative results of two different episodes for navigating to small targets. The left example shows a case in which our GT-Maps successfully navigated to \textit{coca coal glass bottle}, while VLMaps failed. The right example demonstrates a case where both approaches successfully navigate to \textit{remote control}. For each example, the top row shows our semantic map on the left (GT-Map), in which blue points are the navigable points in the scene, the white point is the starting pose, and the yellow star is the selected target. While the VLMaps approach on the right demonstrates the queried LSeg-based map for the target object, the highest activation (\texttt{argmax}) is indicated as a red star. The bottom row shows the last view where the agent stops.   
}
\label{fig:vis_nav}
\end{figure*}

\subsection{Object Goal Navigation}
\label{sect:exp_obj_nav}
% \noindent \textbf{Task Setup:} 
We compare the recently released VLMaps~\cite{vlmaps}  approach on the zero-shot navigation task and compare the performance with the semantic map-based approach generated with our labeling framework, named GT-Maps.
%
% For this experiment, we compare the zero-shot navigation performance using our auto-annotated semantic maps: (1) VLMaps~\cite{vlmaps} and (2) our semantic maps (named \textit{GT-Maps}).
%
% We hypothesize that our auto-annotated semantic maps can be more effective for the semantic-goal navigation task (and more specifically for the long-tail/finer-grained instances) compared to the state-of-the-art zero-shot navigators (using joint embedding space) like VLMaps~\cite{vlmaps}.
% The robot is required to navigate to these four objects sequentially. 
%
The robot observations are RGB, depth, and semantic segmentation images. We also assume the robot can traverse the entire scene. The action space contains moving to a neighboring location on the navigation grid and STOP action. \\
\noindent We adapt \textit{\textbf{VLMaps}} approach to object goal navigation as follows:
\begin{itemize}
    \item Traverse the entire scene, running LSeg~\cite{lseg} and get the pixel embeddings of each frame $I_t$.
 %   \item Initialize the semantic map. The semantic map is a metric map of size $m \times m$. 
    \item Project each frame's pixel embeddings onto the semantic map of size $m \times m$ using the frame's camera pose $(R_t|T_t)$. Take the average of the embeddings projected on the same cell.
    \item Localize the goal object on the semantic map by computing the inner product between the object's text embedding and the semantic map. The cell with the maximum inner product is the location of the object. 
 %   \item Finding the frame pose nearest to the localized to cell.
    \item Run A* algorithm to find the path from the robot's current pose to the selected location.
\end{itemize}
Our (\textit{\textbf{GT-Maps}}) approach is using the semantic map built with auto-labeled semantic segmentations.
For each scene, we select the object goal randomly out of all the object categories detected in the scene.  
We localize the target object in the map and plan the path the same way as the VLMaps approach from several 
randomly sampled start poses. 
% For each AVD scene (20 in total), we first randomly select an object from the existing semantic goals. 
% Then, we sample a random start pose in the scene and compute the shortest path to the target. 
For the \textit{\textbf{VLMaps}}, the target location is obtained by feeding the sampled object target name 
% (which is randomly selected for the \textit{\textbf{GT-Maps}}) 
as input to the LSeg language encoder, query the LSeg heatmap activations and pick the maximum activation of the semantic heatmap as the location of the semantic target. To evaluate the navigation episodes, we manually inspect the STOP location of the agent and its distance to the ground-truth target, both on the top-down semantic map and egocentric RGB views. If the agent stops close to the target, the success rate (SR) would be 1; otherwise, 0. In order to have a more reliable experiment, we repeat the above experiment with three different random seeds (indicated as R1, R2, and R3 on Table~\ref{tab:objgoal}), which means each run covers 20 episodes, and the numbers under each column are the average SR for all 20 episodes for that run. Finally, the last column (Avg-SR) indicates the average and standard deviation of 3 runs for each approach/map.

% \begin{table}[h]
% \caption{Semantic-Goal Navigation evaluation results}
% \label{tab:semgoal}
% \centering
% \begin{tabular}{l|c |c| c| c}
% Dataset & \multicolumn{2}{c}{ADE20K-indoor} & \multicolumn{2}{c}{AVD}\\ 
% \hline
% Method & mIoU ($\%$)  & mIoU-small ($\%$)  & mIoU ($\%$)  & mIoU-small ($\%$)  \\ 
% \hline
% MaskFormer~\cite{Cheng2021PerPixelCI} & 55.7 & 62.3 & 58.6 & 60.2  \\

% MaskFormer+SAM~\cite{Kirillov2023SegmentA} & 56.8 & 63.0 & 64.9 & 70.7 \\ 

% MaskFormer+SAM+Detic (Ours) & \textbf{58.3} & \textbf{63.6}  & 64.2  & 92.8  \\ 

% MaskFormer+SAM+Detic+VG (Ours)  & -- & -- & \textbf{67.4} & \textbf{97.7}  \\ 

% \end{tabular}
% \end{table}

\begin{table}[h]
\caption{Object Goal Navigation Results}
\label{tab:objgoal}
\centering
\begin{tabular}{l|c | c | c | c}
Baselines & R1 & R2 & R3 & Avg-SR ($\%$)\\ 
\hline
VLMaps~\cite{vlmaps} & 84.2 & 73.6 & 78.9 & 78.9$\pm$5.2\\
GT-Maps (Ours) & \textbf{94.7} & \textbf{78.9} & \textbf{94.7} & \textbf{89.4}$\pm$9.1 \\ 
\end{tabular}
\end{table}

As depicted in Table~\ref{tab:objgoal}, we demonstrated that our generated semantic map using our auto-annotation approach could outperform the VLMaps state-of-the-art baseline in zero-shot semantic-goal navigation by $\sim$ 10.5$\%$ in terms of the average success rate. The results of this experiment endorse our hypothesis of the effectiveness of augmenting the semantic maps with the fusion of dense predictions using large-scale pre-trained models through labeling, compared to using large vision-language models in a zero-shot setting. Two qualitative results can be seen in Fig.~\ref{fig:vis_nav}.

\section{Conclusion}
This paper proposes a labeling approach for image semantic segmentation using pre-trained vision models.
We design two downstream tasks based on the segmentation results for object part segmentation and zero-shot robot semantic-goal navigation.
We use the labeling approach to annotate an RGB-D dataset, AVD.
Our experiments demonstrate that:
(i) SAM effectively discriminates object boundaries, especially for small objects and
(ii) The built top-down-view semantic map with our labeling approach for semantic segmentation is competitive in zero-shot semantic-goal navigation compared to VLMaps. 
In terms of future works, our pseudo-labels can be utilized for curating benchmarks for various vision-and-language tasks in indoor scenes, such as grounding (particularly for fine-grained concepts)\cite{glip}, spatial relationships understanding\cite{vsr}, referring expression comprehension\cite{fiber}, descriptive image captioning, instruction following\cite{vlnce}, and beyond.

\section{Acknowledgements}
This work is supported by National Science Foundation, under grant IIS 1925231 NSF NRI. Also, most of the experiments were run on ARGO, a research computing cluster provided by the Office of Research Computing at George Mason University.

%%%%%%%%% REFERENCES
{\small
\bibliographystyle{ieee_fullname}
\bibliography{egbib}
}

\appendix
\vspace{1cm}
\hspace{1cm}\textbf{\Large Supplementary Material}
\vspace{0.5cm}
\section {Experiments}
\subsection{Auto-Labeling}
\begin{figure*}[ht!]
\centering
\includegraphics[width=0.9\textwidth]{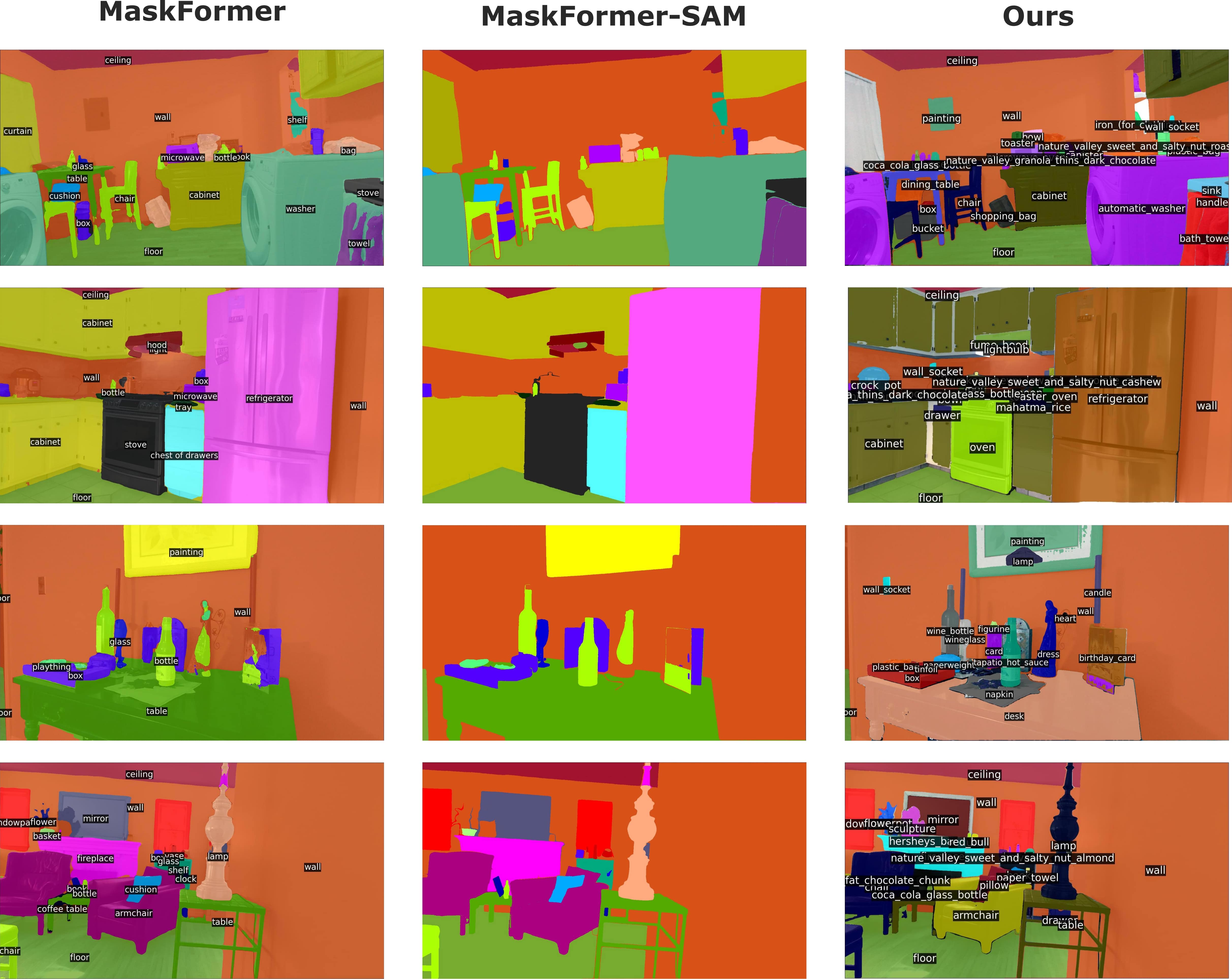}
\caption{
\textbf{Comparison of auto-labeling results on MaskFormer and MaskFormer+SAM, and MaskFormer+SAM+Detic+MV (ours) on AVD images.}
Notably, MaskFormer+SAM demonstrates improved boundary delineation between distinct objects. 
Our approach also utilizes Detic and existing bounding boxes from AVD for getting more object masks, and the multi-view verification stage improves over single-view results.}
\label{fig:vis_avd}
\end{figure*}

In Fig. \ref{fig:vis_avd}, we visualize the results from the baseline methods MaskFormer and MaskFormer+SAM alongside our approach, MaskFormer+SAM+Detic+MV, when applied to the AVD dataset. 

\begin{figure*}[ht!]
\centering
\includegraphics[width=0.9\textwidth]{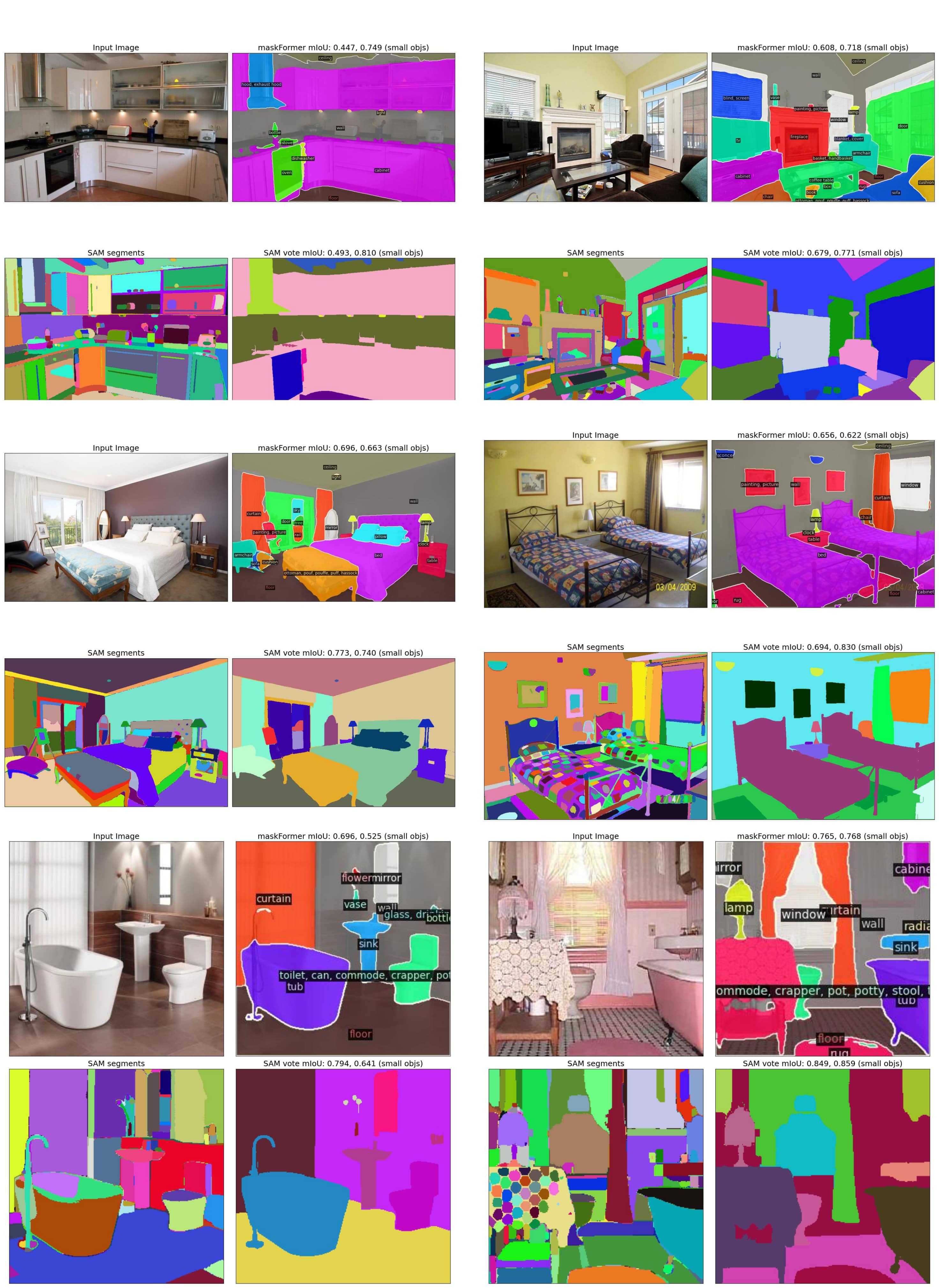}
\caption{
\textbf{A comparison between MaskFormer and MaskFormer-SAM on the ADE20k-indoor dataset.}
In certain examples, MaskFormer-SAM showcases notable improvements, surpassing MaskFormer by $10\%$ in mIoU and $12\%$ in mIoU-small.
}
\label{fig:vis_ade20k}
\end{figure*}

In Fig. \ref{fig:vis_ade20k}, we show the results on the ADE20K dataset. The multi-view verification stage is omitted because of the lack of information to associate different images in the dataset, so only MaskFormer and MaskFormer+SAM results are visualized.

\subsection{Object Part Discovery}
We additionally annotate 'bottle cap' via the object part discovery approach detailed in Sec.5.1.

\begin{figure*}[ht!]
\centering
\includegraphics[width=0.95\textwidth]{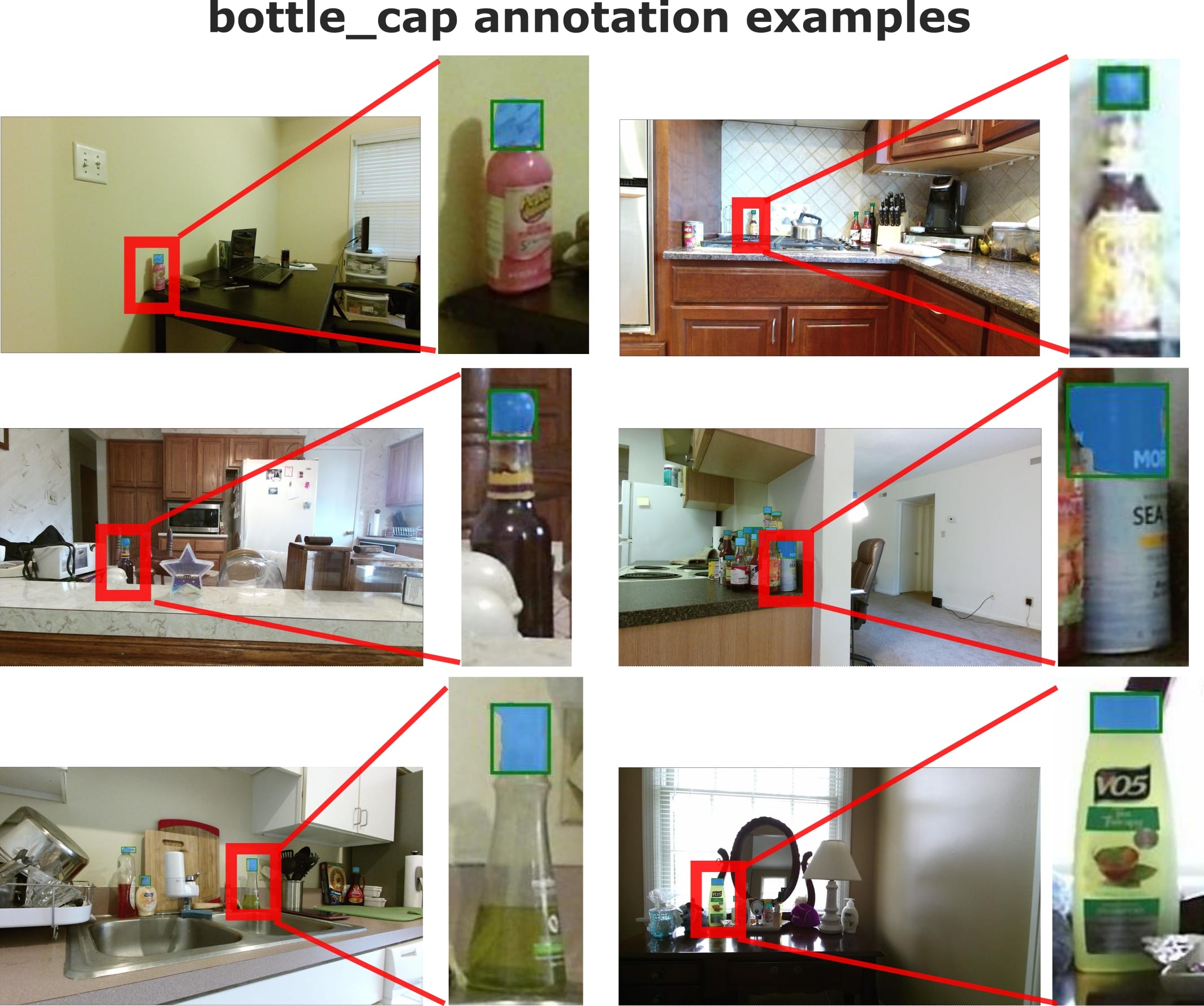}
\caption{
\textbf{This figure gives examples of bottle caps marked as red rectangles.}
For enhanced clarity, the labeled bottle caps have been enlarged to improve visualization.
}
\label{fig:vis_bottle_cap}
\end{figure*}

Fig.~\ref{fig:vis_bottle_cap} shows the annotated bottle caps.

%%%%%%%%% BODY TEXT
\subsection{Object Goal Navigation}
\label{sec:objgoalnavres}

Table.~\ref{tab:objgoalnavrestab} provides comprehensive details about the testing episodes, such as the testing environment, the target object, and the navigation success rates for both the VL-Map and GT-Map approaches.

\begin{figure*}[ht!]
\centering
\includegraphics[width=0.95\textwidth]{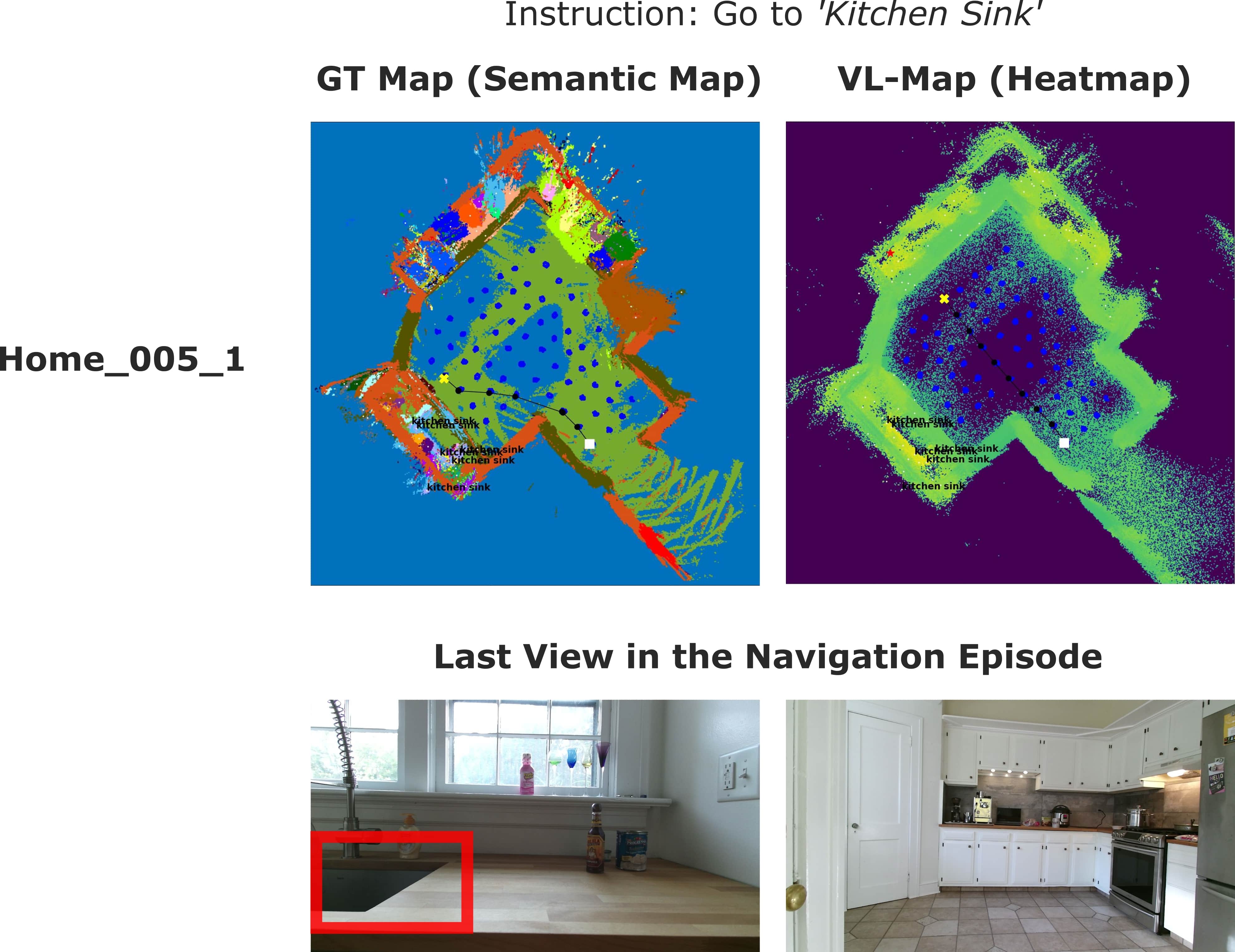}
\caption{
A qualitative result of an episode for navigating to a \textbf{kitchen sink} in \textbf{Home-005}. 
Our approach successfully drives the agent to the sink, while the VL-Map approach fails to approach the target object.
}
\label{fig:vis_nav_005}
\end{figure*}

\begin{figure*}[ht!]
\centering
\includegraphics[width=0.95\textwidth]{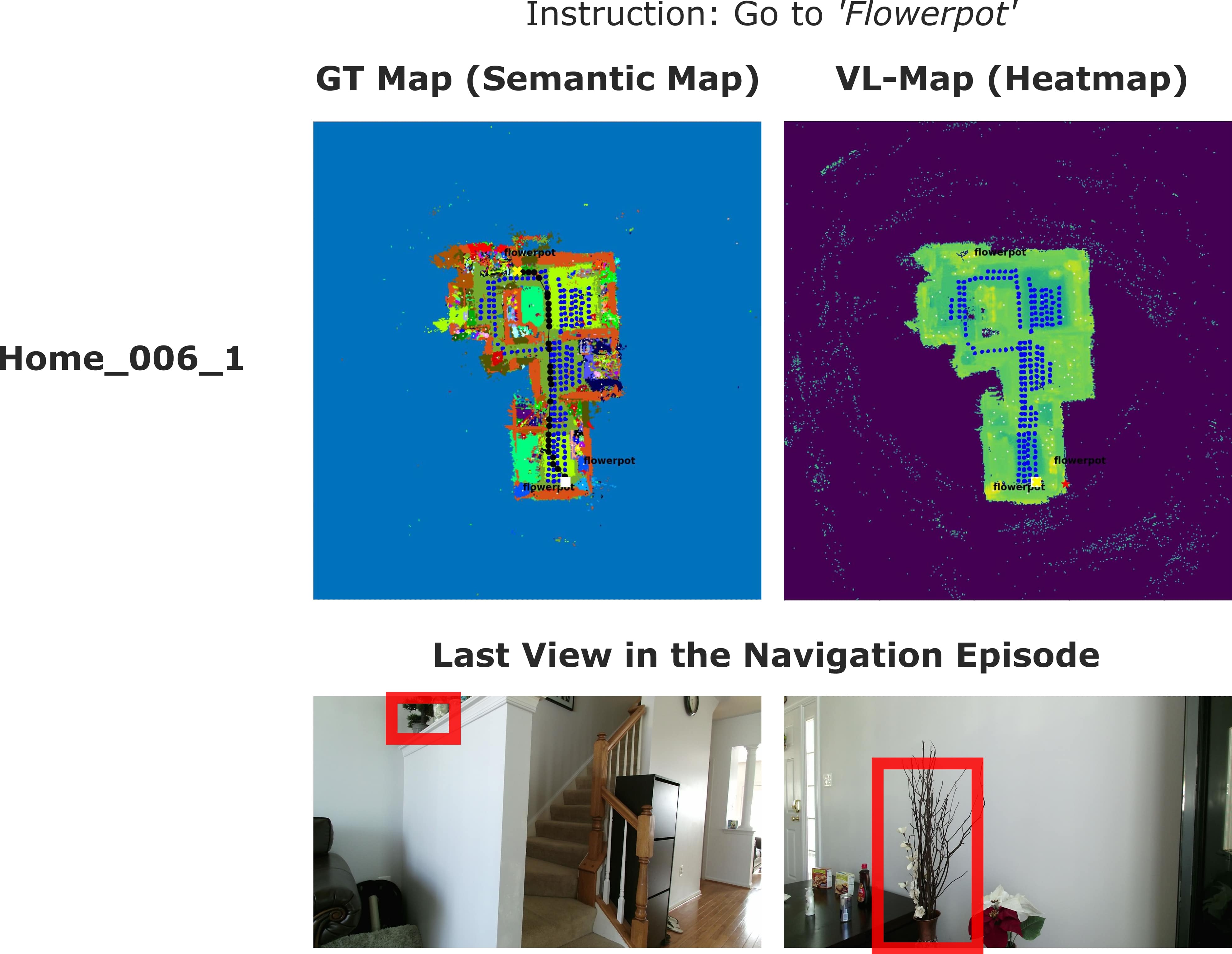}
\caption{
A qualitative result of an episode for navigating to a \textbf{flowerpot} in \textbf{Home-006}.
Both approaches successfully reach the target object.
}
\label{fig:vis_nav_006}
\end{figure*}

\begin{figure*}[ht!]
\centering
\includegraphics[width=0.95\textwidth]{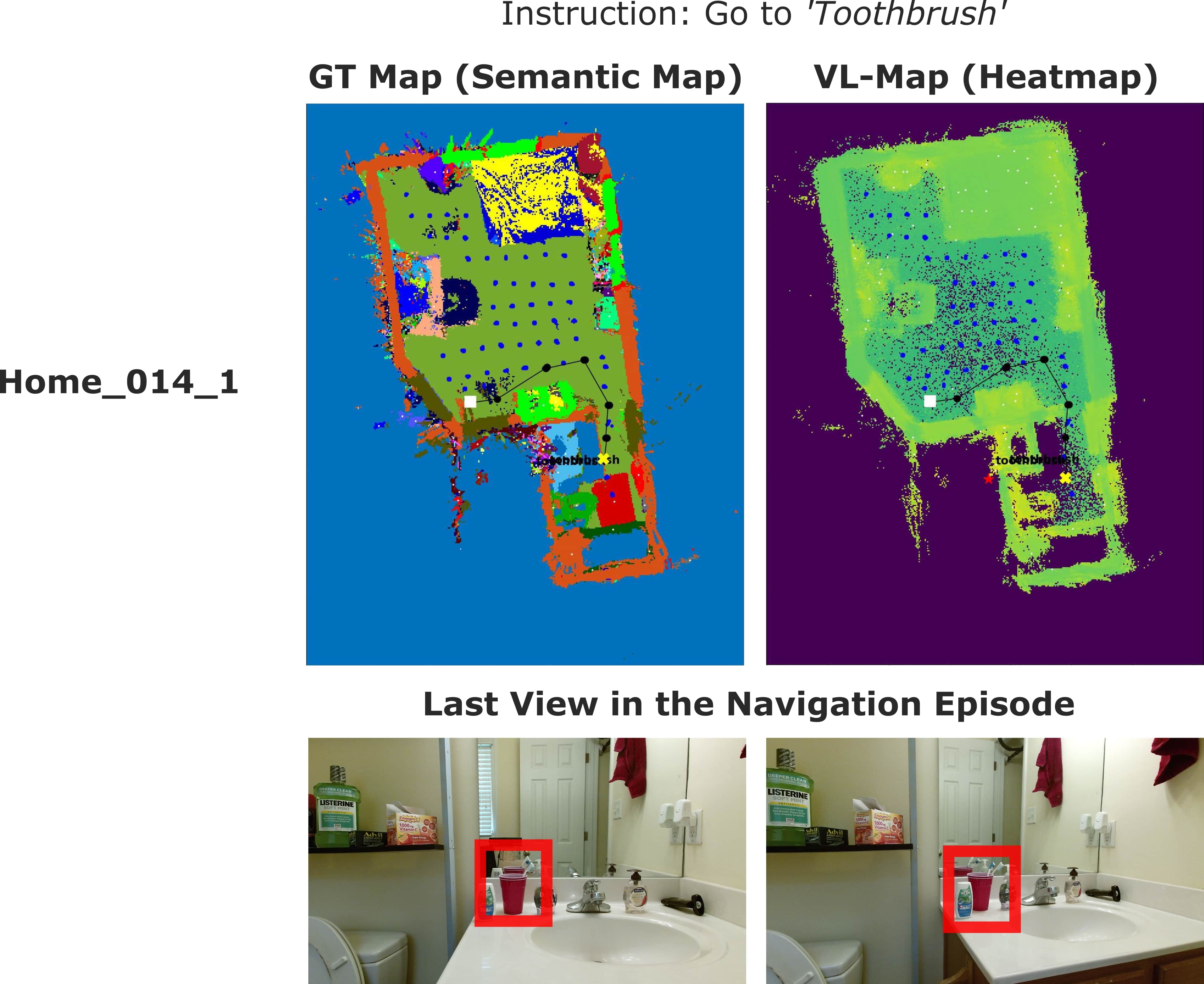}
\caption{
A qualitative result of an episode for navigating to a \textbf{toothbrush} in \textbf{Home-014}. 
Both approaches successfully reach the target object.
}
\label{fig:vis_nav_014}
\end{figure*}

\begin{figure*}[ht]
\centering
\includegraphics[width=0.95\textwidth]{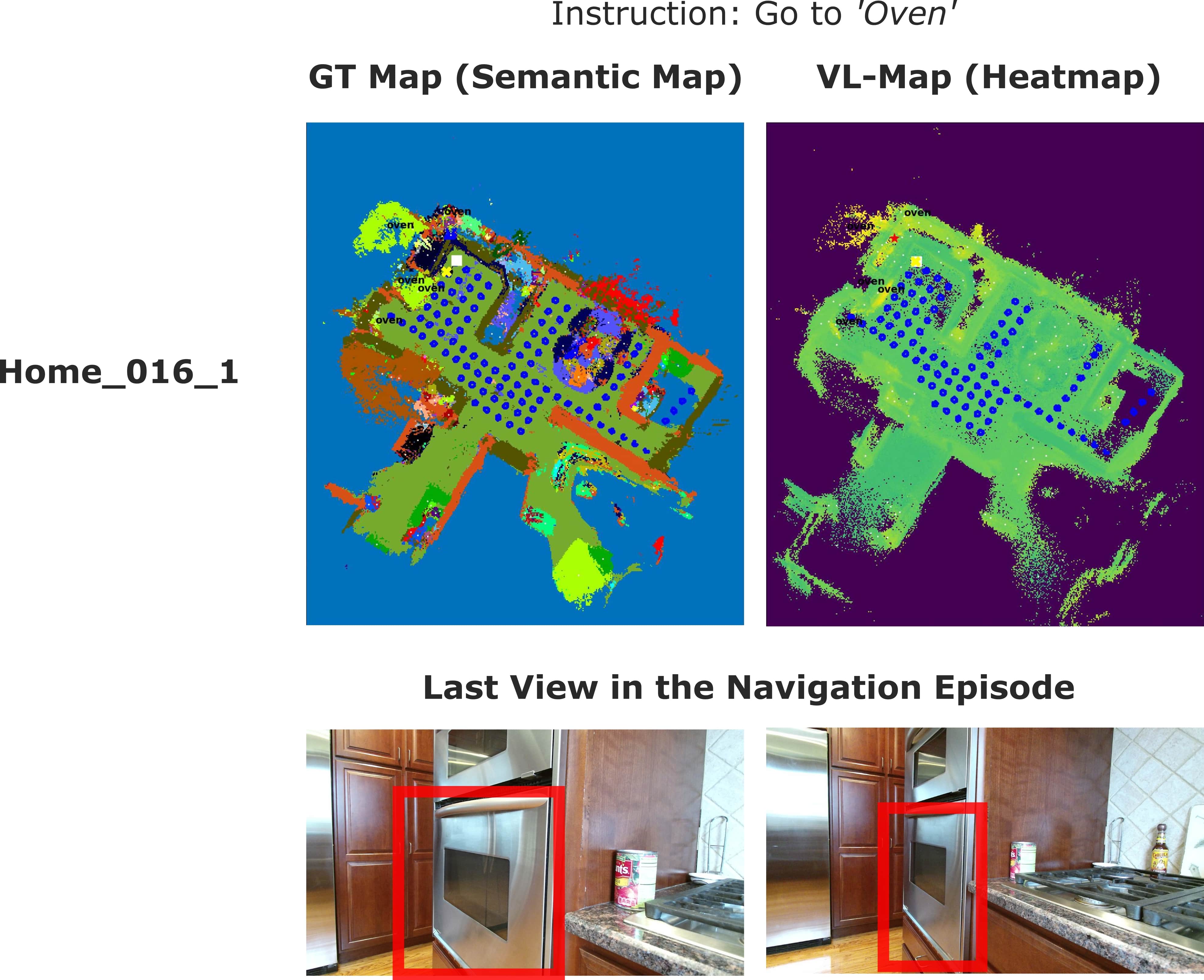}
\caption{
A qualitative result of an episode for navigating to an \textbf{oven} in \textbf{Home-016}. 
Both approaches successfully reach the target object.
}
\label{fig:vis_nav_016}
\end{figure*}

To gain further insight into the performance of our navigation episodes, Fig.~\ref{fig:vis_nav_005}, Fig.~\ref{fig:vis_nav_006}, Fig.~\ref{fig:vis_nav_014} and Fig.~\ref{fig:vis_nav_016} present qualitative results across various AVD environments.

% \begin{table*}[h]
\begin{sidewaystable*}[h]
% \scriptsize
\small
\caption{Detailed results for Object Goal Navigation experiment}
\label{tab:objgoalnavrestab}
\centering
\begin{tabular}{l|c|c|c|c|c|c|c|c|c}
 \multicolumn{3}{c}{R1} & \multicolumn{3}{c}{R2} & \multicolumn{3}{c}{R3} \\
\hline
Run \# & Target & VLMaps & GT-Maps & Target & VLMaps & GT-Maps & Target & VLMaps & GT-Maps \\
\hline
Home-001-1 & remote control & 1 & 1 & sink & 1 & 1 & sofa & 1 & 1 \\
Home-001-2 & sponge & 0 & 1 & sofa bed & 1 & 1 & cabinet & 1 & 1 \\
Home-002-1 & sofa & 1 & 1 & handbag & 0 & 0 & sofa & 1 & 1 \\
Home-003-1 & dining table & 1 & 1 & toilet & 1 & 0 & fireplace & 1 & 1 \\
Home-003-2 & cabinet & 1 & 1 & bath towel & 1 & 1 & cabinet & 1 & 1 \\
Home-004-1 & refrigerator & 1 & 1 & dining table & 1 & 1 & blanket & 0 & 1 \\
Home-004-2 & sofa & 1 & 1 & crystal hot sauce & 1 & 1 & plastic bag & 1 & 1 \\
Home-005-1 & kitchen sink & 0 & 1 & refrigerator & 1 & 1 & coffee maker & 1 & 1 \\
Home-005-2 & bottle & 1 & 1 & kitchen sink & 1 & 1 & faucet & 0 & 1 \\
Home-006-1 & bath mat & 1 & 1 & box & 1 & 1 & flowerpot & 1 & 1 \\
Home-007-1 & chair & 1 & 1 & chair & 1 & 1 & sofa & 1 & 1 \\
Home-008-1 & coca cola glass bottle & 0 & 1 & cushion & 1 & 1 & refrigerator & 0 & 0 \\
Home-010-1 & sofa bed & 1 & 1 & jacket & 0 & 1 & box & 1 & 1 \\
Home-011-1 & hand towel & 1 & 0 & oven & 1 & 1 & nature valley sweet and salty nut cashew & 1 & 1 \\
Home-013-1 & pringles bbq & 1 & 1 & box & 0 & 1 & refrigerator & 1 & 1 \\
Home-014-1 & toothbrush & 1 & 1 & book & 1 & 1 & chair & 1 & 1 \\
Home-014-2 & desk & 1 & 1 & suitcase & 0 & 0 & drawer & 0 & 1 \\
Home-015-1 & bottle & 1 & 1 & easel & 0 & 0 & dresser & 1 & 1 \\
Home-016-1 & cabinet & 1 & 1 & oven & 1 & 1 & teakettle & 1 & 1 \\
\hline
Average & -- & 84.2 & 94.7 & -- & 73.6 & 78.9 & -- & 78.9 & 94.7 \\
\end{tabular}
% \end{table*}
\end{sidewaystable*}

\end{document}